\documentclass{article}
\usepackage{PRIMEarxiv}
\usepackage[utf8]{inputenc} 
\usepackage[T1]{fontenc}    
\usepackage{hyperref}       
\usepackage{url}            
\usepackage{booktabs}       
\usepackage{amsfonts}       
\usepackage{nicefrac}       
\usepackage{microtype}      
\usepackage{lipsum}
\usepackage{fancyhdr}       
\usepackage{graphicx}       
\usepackage{amsmath}       
\graphicspath{{media/}}     
\usepackage{algorithm}
\usepackage{multirow}
\usepackage{algpseudocode}
\title{DAPA: \underline{D}istribution \underline{A}ware \underline{P}iecewise \underline{A}ctivation Functions for On-Device Transformer Inference and Training
\thanks{ 
\textbf{Accepted for publication at the 63rd IEEE/ACM Design Automation Conference (DAC 2026)}} 
}

\author{
  Maoyang Xiang, Bo Wang \\
  Information Systems Technology and Design \\
  Singapore University of Technology and Design \\
  Singapore\\
  \texttt{\{maoyang\_xiang, bo\_wang\}@sutd.edu.sg} \\
}

\begin{document}
\maketitle

\begin{abstract}
Non-linear activation functions play a pivotal role in on-device inference and training, as they not only consume substantial hardware resources but also impose a significant impact on system performance and energy efficiency.
In this work, we propose Distribution-Aware Piecewise Activation (DAPA), a differentiable and hardware-friendly activation function for Transformer architectures by exploiting the distribution of pre-activation data. DAPA employs a non-uniform piecewise approximation that allocates finer segments to high-probability regions of the distribution, improving generalizability over prior piecewise linear methods. The resulting approximation is further quantized using Distribution-Weighted Mean Square Error to reduce latency and resource utilization for hardware deployment.
Our HLS implementation demonstrates that DAPA speeds up GELU computation by 16$\times$ and decreases DSP utilization by 16$\times$
while maintaining comparable or better performance across vision Transformers and GPT-2 models.
\end{abstract}

\keywords{Activation Function \and Distribution \and Piecewise Linear Approximation}

\section{Introduction}
Activation functions are fundamental components of modern Deep Neural Networks (DNNs) as they introduce essential non-linearity for models to learn complex data patterns in the real world. 
To enable efficient DNN acceleration at the edge, approximated activation functions are often adopted. As a result, the quality and efficiency of these approximations become crucial for overall system performance and energy efficiency. \\
In particular, activation function approximation becomes a major bottleneck to accelerating Transformer architectures. Although matrix multiplications in Transformers are highly parallel, overall throughput is often constrained by the latency of non-linear activation functions \cite{kim2021bert}. However, approximating these functions typically involves computationally expensive operations such as higher-order polynomials that are inherently slower than simple arithmetic, which can stall parallel processing pipelines. \\ 
Consequently, how to efficiently implement these non-linear operations becomes critical for enhancing the overall efficiency of on-device Transformer execution.
To handle this challenge, strategies such as Look-Up Tables (LUTs), polynomial approximations, and piecewise linear functions have been proposed. They are all designed to minimize the Mean Squared Error (MSE) with respect to the original activation function for approximation. 
However, a smaller MSE does not necessarily translate to better model performance when utilizing the approximated functions. This is because MSE implicitly assigns uniform weighting over all pre-activation inputs, regardless of the actual data distribution.
Consequently, approximations optimized for minimizing MSE may not generalize well and can lead to inefficient hardware resource utilization as unnecessary precision can be allocated to statistically insignificant regions. \\
To address this limitation, we propose Distribution-Aware Piecewise Linear Approximation (DAPA) functions, a differentiable activation approximation for on-chip Transformer inference and training. Our approach leverages the actual input data distribution, allowing the approximation to focus on high-probability regions that pose a significant impact on performance. This leads to a more accurate and hardware-efficient approximation with improved generalizability for various Transformer models. The main contributions of this paper are summarized as follows:
\begin{itemize}
\item We propose a novel approach that exploits the input probability density function to approximate both activation functions and their derivatives, which can be generalized to a variety of Vision Transformer and the GPT-2 models. 
Our evaluation demonstrates that it speeds up GELU computation by $16\times$ without compromising accuracy. 
\item We introduce Distribution-Weighted Mean Squared Error (DWMSE) as a new loss function for approximation. Our experiment shows that DWMSE exhibits a stronger correlation with model performance variations than conventional MSE, leading to improved network performance with the resulting approximation. 
\item We propose a 16-bit fixed-point (Fix16) quantization scheme for DAPA that automatically selects integer and fractional precision under a DWMSE-guide error budget, achieving inference accuracy comparable to a non-quantized baseline.
\item We demonstrate a 16$\times$ reduction in DSP utilization for GELU and 48$\times$ reduction for Softmax, both with significant savings in flip-flops and LUT resources compared to prior Fix16 implementations.  
\item We show that DAPA-based GELU functions can be successfully trained from scratch and converge at the same rate as standard GELU, while achieving slightly higher accuracy in Vision Transformer models. 
\end{itemize}

The remainder of this paper is organized as follows: Section 2 reviews prior work on activation function approximation.
Section 3 details the proposed DWMSE metric and the DAPA approach, while Section 4 presents the experiential results of DAPA across multiple tasks. 
Finally, Section 5 concludes this work. Our source code is available at \url{https://github.com/MayerUser/DAPA_Activation}

\section{Related Work}
\label{sec:RELAT}
The complexity of modern activation functions presents a significant challenge for efficient hardware implementation \cite{park2024implementation}. In particular, their non-linear operations, such as exponentials, are prohibitively expensive in terms of area, latency, and power, motivating the use of approximation techniques \cite{ peltekis2024reusing}. Among existing approaches, LUTs provide simplicity and relatively low latency, but their memory requirements grow exponentially with precision \cite{li2022auto, bouraya2024comparative}. Algorithmic approximations present an alternative. For instance, Taylor series-based methods can achieve high accuracy but require a substantial number of Multiply-Accumulate (MAC) operations \cite{guo2024hg, bouraya2024comparative, lee2003some, temurtas2004study}. In contrast, the CORDIC algorithm offers a more area-efficient solution by solely using shifts and additions, though its iterative nature results in higher latency \cite{guo2024hg, kokane2025vinci}. 

The limitations of these methods lead to the rise of hybrid piecewise approximation. This technique partitions the function into multiple segments and approximates each using a simple linear model or a low-degree polynomial \cite{li2022auto, reggiani2023flex, sadeghi2024peano}. By localizing the approximation, it achieves higher accuracy with minimal hardware cost. Specifically, rather than storing full-precision function values, a piecewise approximation requires only a small LUT for segment coefficients and a MAC unit to reconstruct the function output. This structure substantially reduces both memory footprint and logic utilization, providing an effective balance between accuracy and hardware efficiency \cite{huang2025ispa, li2022auto, peltekis2024reusing}.

The piecewise approximation method has been widely adopted and evolved into multiple variants. For instance, the Internal Symmetry Piecewise Approximation (ISPA)~\cite{huang2025ispa} leverages the odd symmetry of the internal Gaussian error function to approximate GELU, effectively halving the size of the coefficient memory. \cite{huang2023hardware} has proposed a hardware-efficient piecewise linear activation function as an alternative to GELU. However, the performance is constrained by the configurability of the parameters, limiting the generalizability across different models and tasks.
At the architectural level, the Flex-SFU accelerator~\cite{reggiani2023flex} employs non-uniform segmentation over a fixed input range, assigning finer intervals to high-curvature regions. Consequently, it achieves higher accuracy than uniform schemes. However, for a given function and number of segments, these intervals are statically calculated offline and remain unchanged at run time, without adapting to input-dependent data distributions.

It is worth noting that most of these approaches merely focus on minimizing function-level error (e.g., MSE) between the approximated and the original activations~\cite{marchisio2023swifttron}. 
Although effective in some cases, this methodology is inherently data-agnostic. It treats all parts of the input range as equally important and ignores the actual distribution of pre-activation values. 
As a result, hardware resources are often misallocated, providing unnecessary precision to rarely visited regions while underserving high-probability inputs. This limitation motivates a shift from conventional function-level error minimization to distribution-aware error minimization. 

\section{Methodology of DAPA}
In this section, we introduce a distribution-aware error metric, Distribution-Weighted Mean Squared Error (DWMSE), together with a Distribution-Aware Piecewise Activation (DAPA) for on-device Transformer inference and training. 

\subsection{Distribution-Weighted Mean Squared Error} 
\begin{figure}
  \centering
  \includegraphics[width=0.6\textwidth]{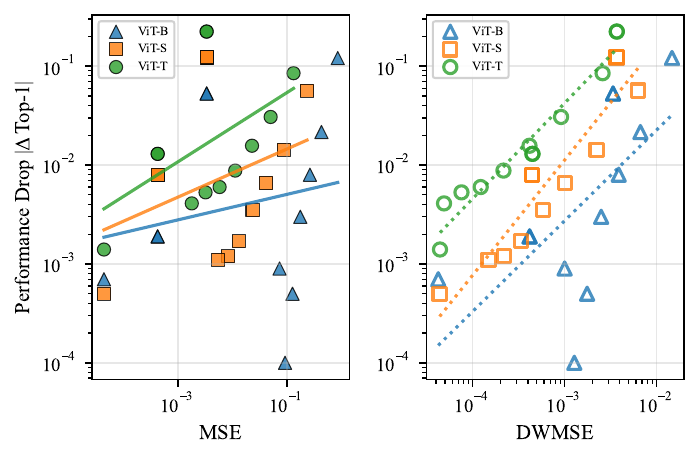}
  \caption{  The relationship between MSE/DWMSE and the difference in Top-1 ImageNet-1K classification accuracy relative to the FP32 baseline, evaluated across three Vision Transformer variants.}
  \label{fig:DWMSE_ACC}
\end{figure}

When designing hardware-friendly approximations, MSE is a commonly used error function to guide approximation optimization. However, we observe a remarkable discrepancy where approximations with low MSE values can still lead to substantial degradation in model accuracy, as Fig.\ref{fig:DWMSE_ACC} illustrates. This indicates that MSE, by assigning equal importance to all input ranges, can not precisely capture the components most critical to network performance.

\begin{figure}
  \centering
  \includegraphics[width=0.6\textwidth]{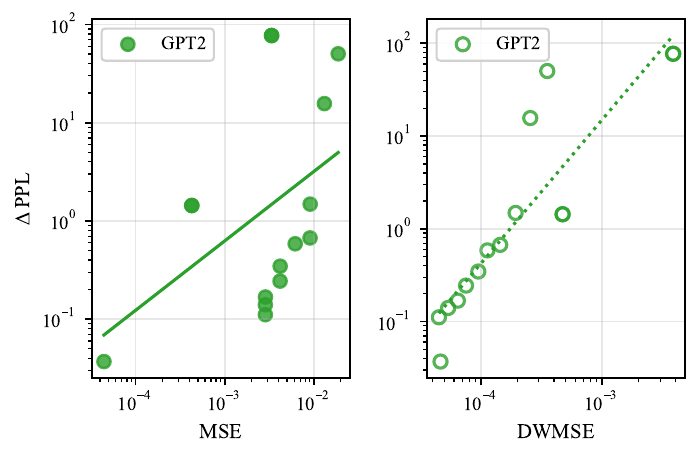}
  \caption{  The relationship between MSE/DWMSE and the difference in WikiText2 PPL relative to the FP32 baseline, evaluated on GPT2.}
  \label{fig:DWMSE_PPL}
\end{figure}

Specifically, the inputs to an activation function in a well-trained Transformer model typically exhibit a non-uniform probability distribution. 
However, the intrinsic distribution characteristics are ignored by MSE, which penalizes errors in low-probability and high-probability regions equally.
To address this limitation, we propose DWMSE as a new metric for evaluating performance variation arising from approximated activation functions.
It extends the standard MSE by incorporating the probability density function (PDF) of the input data, $p(x)$. The DWMSE within interval $[a,b]$ is defined as:

\begin{equation}
    \text{DWMSE} = \frac{1}{b-a}\int_{a}^{b} p(x) (\sigma(x) - \hat{\sigma}(x))^2 \,dx    
\end{equation}

where $\sigma(x)$ denotes the original activation function, $\hat{\sigma}(x)$ denotes its approximation.
The formula weights the squared error according to its probability of occurrence, highlighting the most frequently encountered inputs. 

\begin{table}[htbp]
 
\caption{Comparison of correlation coefficients and 95\% Fisher CIs.}
\begin{center}
\begin{tabular}{llcccc}
\toprule
\multirow{2}{*}{\textbf{Model}} & \multirow{2}{*}{\textbf{Metric}} & \multicolumn{3}{c}{\textbf{Correlation Coefficients}} & \multirow{2}{*}{\textbf{Fisher CI}} \\
\cline{3-5} 
& & \textbf{\textit{ r}} & \textbf{\textit{ $\rho$}} & \textbf{\textit{ $\tau$}} & \\
\midrule
\multirow{2}{*}{\textbf{ViT-Base}} & MSE & 0.191 & 0.359 & 0.312 & [-0.43, 0.69] \\
& DWMSE & 0.646 & 0.754 & 0.562 & [0.12, 0.89] \\
\midrule
\multirow{2}{*}{\textbf{ViT-Small}} & MSE & 0.323 & 0.169 & 0.250 & [-0.31, 0.76] \\
& DWMSE & 0.938 & 0.923 & 0.812 & [0.79, 0.98] \\
\midrule
\multirow{2}{*}{\textbf{ViT-Tiny}} & MSE & 0.490 & 0.528 & 0.438 & [-0.12, 0.83] \\
& DWMSE & 0.973 & 0.979 & 0.938 & [0.90, 0.99] \\
\midrule
\multirow{2}{*}{\textbf{GPT-2}} & MSE & 0.230 & 0.490 & 0.449 & [-0.32, 0.66] \\
& DWMSE & 0.903 & 0.943 & 0.864 & [0.73, 0.97] \\
\bottomrule
\end{tabular}
\label{tab:Correlation_DWMSE}
\end{center}
\end{table}

To validate the efficacy, we investigate the model performance changes against MSE and DWMSE metrics, respectively, and analyze how approximation error correlates with accuracy degradation. We study this using ViT-Base/Small/Tiny\cite{dosovitskiy2020image,rw2019timm} models on the ImageNet-1K dataset\cite{imagenet15russakovsky}, and the GPT-2 Base model\cite{radford2019language} on the WikiText-2 dataset.
All MSE and DWMSE values are computed over the range of $[-4,4]$, following a common approach~\cite{sadeghi2024peano}.
As Fig.~\ref{fig:DWMSE_ACC} shows, DWMSE exhibits a more linear correlation with the accuracy change compared to MSE. It is worth noting that the correlation for ViT-Base is weaker than for the other models, primarily because many ViT-Base cases exhibit a tiny accuracy drop.
Subsequently, we use Pearson coefficient $r$, Spearman coefficient $\rho$, Kendall coefficient $\tau$, and their 95\% Fisher Confidence Intervals (CIs) to further quantify the correlation between the error metrics and performance changes. 

Table~\ref{tab:Correlation_DWMSE} shows that for all ViT models, DWMSE exhibits a consistently higher correlation with the performance change compared to MSE. Moreover, the $95\%$ Fisher CIs for DWMSE are always narrower than those for MSE, indicating a more precise estimation across all tested models.

For the GPT-2 Base model, DWMSE also exhibits higher coefficient values with respect to the perplexity change ($\Delta\text{PPL}$), indicating a stronger correlation. As further illustrated in Fig.~\ref{fig:DWMSE_PPL}, the $\Delta\text{PPL}$ shows a more linear relationship with DWMSE, suggesting that DWMSE provides a more accurate measure to capture performance variations. 

\begin{figure}
  \centering
  \includegraphics[width=0.6\textwidth]{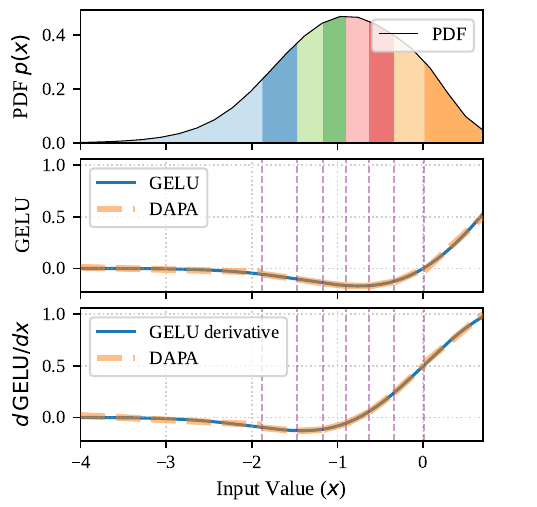}
  \caption{  The DAPA method approximates the GELU function and its derivative by partitioning the probability density function (PDF) into $N$ quantile regions.}
  \label{fig:dapa}
\end{figure}

\subsection{Design of Distribution-Aware Piecewise Activation}

The core principle of our DAPA function is to achieve a highly accurate approximation by allocating precision according to the input data's probability distribution. Rather than dividing the input range into uniform segments, DAPA partitions the cumulative probability into $N$ equal segments, ensuring that each segment represents an equal portion (i.e., $1/N$) of the probability mass, as Figure \ref{fig:dapa} illustrates. As a result, this scheme generates finer granularity for pre-activation data with high probability density and coarser granularity for data with low probability density. The boundaries of these segments, or "knots," are determined using the inverse of the Cumulative Distribution Function (CDF), $F^{-1}(x)$. The CDF, $F(x)$, is defined as the integral of the probability density function $p(t)$: 
\begin{equation}
    F(x) = \int_{-\infty}^{x} p(t) \,dt  
\end{equation}

where $t$ denotes the pre-activation input of the non-linear function. In practice, $p(t)$ is captured from the actual distribution of pre-activation values collected by running the network on $M$ real samples (images or tokens) and aggregating all occurrences across layers.

Thus the $n$-th knot, $k_i$, is calculated as: 

\begin{equation}
    k_n = F^{-1}\left(\frac{n}{N}\right) \quad \text{for } n = 1, \dots, N-1 
\end{equation}

Once the knots $[k_0, k_1, \dots, k_{N-1}]$ are determined, we design the optimal linear approximation function, $\hat{\sigma}(x)$, that best approximates the activation function, $\sigma(x)$ within each interval $[k_n, k_{n+1}]$. The coefficients $a_n$ and $b_n$ for the linear approximation of the targeted activation function in the $n$-th segment are obtained by solving the following optimization problem: 

\begin{equation}
\label{equ:DWMSE}
\begin{split}
\{a_n, b_n\} &= \underset{a, b}{\arg\min} \text{DWMSE}(\sigma(x),\hat{\sigma}(x))\bigg|_{k_{n+1}}^{k_n}\\
\end{split}
\end{equation}

The continuous optimization problem can be discretized and solved efficiently as a Weighted Least Squares (WLS) problem\cite{kiers1997weighted}. The integral is approximated by a weighted summation over $m$ samples, $x_i$, drawn from the distribution $p(x)$ within the interval $[k_n, k_{n+1}]$. Therefore, the objective function $L$ to be minimized is:

\begin{equation}
    \underset{a, b}{\arg\min} \, L(a, b) = \sum_{i=1}^{m} p_i (\sigma(x_i) - (ax_i + b))^2
\end{equation}

where $p_i = p(x_i)$ is the corresponding weight (probability) for each sample.

As a key activation in Transformer architectures, \textit{softmax} differs from element-wise nonlinearities such as GELU in that it performs exponentiation and normalization over a vector. We adopt the standard shifted exp-sum formulation

\begin{equation}
    \mathrm{softmax}(x_i) = \frac{\exp(x_i - x_{\max})}{\sum_{j=1}^{N} \exp(x_j - x_{\max})}
\end{equation}

which guarantees that all exponential inputs are non-positive and the outputs are within the range of $(0,1]$, preventing hardware overflow. In particular, DAPA is designed to compute the exponential operations in the softmax function, while the remaining computations are offloaded to a MAC unit for efficiency.


Similarly, the linear approximation of the derivative of the activation function, $\sigma'(x)$, within each interval $[k_s, k_{s+1}]$, can be defined as:
\begin{equation}
 \{\bar{a}_n, \bar{b}_n\} = \underset{\bar{a}, \bar{b}}{\arg\min} \int_{k_n}^{k_{n+1}} p(x) (\sigma'(x) - (\bar{a}x + \bar{b}))^2 \,dx
\end{equation}

which can be seamlessly integrated into the backpropagation process for training networks from scratch.

\begin{algorithm}[t]
 
\caption{ Fixed-Point Format Selection Guided by DWMSE}
\label{alg:dapa_quant}
\begin{algorithmic}[1]
\Require $\hat{\sigma}^{\mathrm{fp32}}(x)$ full precision DAPA, input range $[a,b]$, scaling factor $\theta$, $\mathrm{BIT\_MAX}=16$
\Ensure $Q(m,n)$
\State $\mathrm{threshold} \gets \theta \cdot \mathrm{DWMSE}(\hat{\sigma}^{fp32})$
\State $m \gets \lceil \log_2(\max(|a|,|b|)) \rceil + 1$
\State $n \gets 0$
\While{$m + n < \mathrm{BIT\_MAX}$ \textbf{and} $\mathrm{DWMSE}(\hat{\sigma}^{Q(m,n)}) > \mathrm{threshold}$}
    \State $n \gets n + 1$
\EndWhile
\State \Return $(m,n)$
\end{algorithmic}
\end{algorithm}

Benefiting from DWMSE, we further quantize DAPA to a fixed-point representation for the inference stage. We first compute the DWMSE of DAPA under floating-point arithmetic. Then, we introduce a scaling factor $\theta$ and define the admissible error threshold as $\theta \times \text{DWMSE}$. Next, we determine the number of integer bits from the maximum absolute input value of the activation function from its distribution, and iteratively increase the number of fractional bits while recomputing DWMSE until it falls below the threshold. In this work, we constrain the total bit-width to be no greater than 16 so that the subsequent hardware design can uniformly adopt a 16-bit fixed-point format.

\section{Experimental Results} \label{sec:results} 
In this section, we present a comprehensive evaluation of DAPA, including network performance results and HLS-based hardware implementation. We also report the training performance of ViT models with DAPA functions.
\subsection{Impact of Number of Input Samples}

Since DAPA is based on real data distributions, determining the appropriate number of input samples to the network to obtain a reliable distribution is critical. We thus design an experiment to investigate the relationship between the number of samples $M$ used to generate DAPA and the resulting model accuracy. Using ViT-small, we randomly sample varying numbers of images from the ImageNet dataset to construct DAPA functions with different segment counts. The pretrained ViT-Small model with these DAPA functions is then evaluated to measure classification accuracy. 

\begin{figure}[t]
  \centering
  \includegraphics[width=0.60\textwidth]{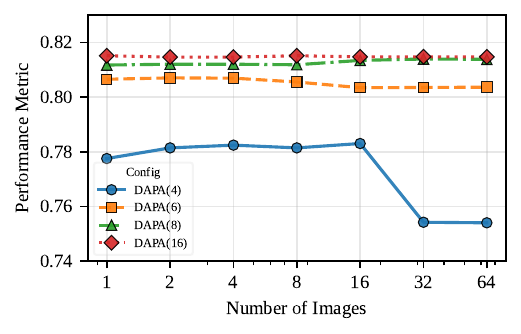}
  \caption{  DAPA functions generated from distributions with different numbers of images are evaluated on the ViT-Small.}
  \label{fig:distributionSamples}
\end{figure}

As Figure \ref{fig:distributionSamples} illustrates, the accuracy remains with very small variations and is largely insensitive to the number of input samples used for distribution modeling. In particular, for the configurations with more than 4 segments, the final accuracy varies by less than $0.36\%$ across different numbers of input samples.
For the ViT-Small model with DAPA(16), we observe only modest variations where the highest accuracy is 81.51\% and the lowest is 81.46\%, achieving a mean accuracy of 81.47\% with a variance of $4.8\times10^{-8}$.
Overall, DAPA with 16 segments can leverage data distribution for approximation from a small number of input samples.

\subsection{Performance on Image Classification}

\begin{table*}[htbp]
\small
\caption{Performance of Vision and NLP Transformer models with different approximation methods}
\begin{center}
\setlength{\tabcolsep}{2pt}
\begin{tabular}{ccccccccccc}
\toprule
\textbf{Model} & \textbf{Torch, $^{\mathrm{a}}$F} 
& \textbf{$^{\mathrm{b}}$M,$^{\mathrm{c}}$A, F}
& \textbf{$^{\mathrm{d}}$D, A, F} 
& \textbf{D, $^{\mathrm{e}}$S, F} 
& \textbf{D, S\&A, F} 
& \textbf{D, S\&A,$^{\mathrm{f}}$Q} 
& \textbf{\cite{sadeghi2024peano}} 
& \textbf{\cite{marchisio2023swifttron}} 
& \textbf{\cite{wang2024improving}} 
& \textbf{\cite{huang2023hardware}}
\\
\midrule
ViT-Tiny & 75.51\% & 73.82\%  & 75.51\% & 75.44\% &75.42\% & 75.17\% (Q9.7)$^{\mathrm{g}}$ & -& - &- &-\\
ViT-Small & 81.40\% & 80.81\% & 81.41\% & 81.37\%  &81.40\% & 81.30\% (Q8.8) & -& - & - &-\\
ViT-Base & 81.66\% & 81.35\% & 81.69\% & 81.69\% &81.70\% & 81.61\% (Q7.7) & -& - & - &-\\
DeiT-Tiny & 74.53\% & 73.92\% & 74.55\% & 74.47\%  &74.43\% & 74.37\% (Q6.9) & -& - & 71.47\% &-\\
DeiT-Small & 81.17\%& 80.85\% & 81.11\% & 81.21\%  &81.16\% & 81.11\% (Q6.8)& 79.36\%& 79.11\% & 79.44\% & 80.34\%\\
DeiT-Base & 83.32\% & 83.13\% & 82.90\% & 83.28\%  &82.88\% & 82.97\% (Q7.5)& -& - & 82.92\% &-\\
Swin-Small & 83.20\% & 82.96\% & 83.24\% & 83.20\%  &83.23\% & 83.24\% (Q7.5)& -& - & 82.91\% &-\\
Swin-Base & 85.12\% & 84.81\% & 85.08\% & 85.11\%   &85.05\% & 85.05\% (Q7.5)& -& - & 83.36\% &-\\
\midrule
GPT2 (PPL) & 29.37 & 36.50 & 29.47 & 29.37  & 29.47 & 29.47 (Q7.9) & - & - & 31.02 & - \\
\bottomrule
\multicolumn{5}{l}{$^{\mathrm{a}}$\textbf{F}: Data type configured as 32-bit floating point.} &
\multicolumn{6}{l}{$^{\mathrm{b}}$\textbf{M}: The approximation function derived by minimizing MSE.}\\
\multicolumn{5}{l}{$^{\mathrm{c}}$\textbf{A}: DAPA methodology implemented in activation layer.} &
\multicolumn{6}{l}{$^{\mathrm{d}}$\textbf{D}: DAPA employing 16-segment piecewise linear approximation.}\\
\multicolumn{5}{l}{$^{\mathrm{e}}$\textbf{S}: DAPA methodology implemented in softmax layer.} &
\multicolumn{6}{l}{$^{\mathrm{f}}$\textbf{Q}: DAPA quantized using fixed-point number with $\theta=1.05$.} \\
\multicolumn{6}{l}{$^{\mathrm{g}}$ The activation input and output for the DAPA algorithm utilize Q notation.} \\
\end{tabular}
\label{tab:af_comparison}
\end{center}
\end{table*}

We evaluate the performance of the DAPA method with 16 segments, DAPA(16), on a wide range of vision Transformer models, including ViT-Tiny/Small/Base, DeiT-Tiny/Small/Base, and Swin-Small/Base on the ImageNet-1K dataset\cite{dosovitskiy2020image,touvron2021training,liu2022swin}. 
We evaluate DAPA(16) under five configurations, including an MSE baseline where we use MSE as a loss function, DWMSE-guided softmax-only and GELU-only approximations, and a joint softmax+GELU approximation.
As shown in Table~\ref{tab:af_comparison}, the DWMSE-based DAPA approximation for activation layers (i.e., (D, A, F)) attains comparable or slightly higher accuracy than the PyTorch baseline. It also exhibits superior accuracy across almost all the architectures compared to the MSE-based method. The only exception is DeiT-Base, where the MSE-based approach shows a slightly higher accuracy. 
Moreover, when applying DAPA(16) only to softmax (i.e., (D, S, F)), all models achieve comparable accuracy to the PyTorch baseline, where the maximum drop is kept within $0.06\%$. Applying DAPA(16) only to GELU (i.e., (D, A, F)) also preserves performance, with a maximum drop of $0.42\%$ on DeiT-Base. When both softmax and GELU are approximated (i.e., (D, S\&A, F)), the maximum accuracy drop remains as $0.44\%$ on DeiT-Base. Under the DWMSE-guided quantization (i.e., 16-bit fixed-point data (Q9.7)), the additional loss is kept at most $0.23\%$ on ViT-Tiny.

We compare DAPA with state-of-the-art Vision Transformer approximation approaches. 
As Table~\ref{tab:af_comparison} shows, DAPA exceeds the accuracy of the benchmarks such as PEANO-ViT \cite{sadeghi2024peano} and SwiftTron\cite{marchisio2023swifttron}, even after quantization, where our baseline and the prior work's baseline are very close. 
\subsection{Performance on Natural Language Processing Tasks}
We also evaluate DAPA(16) in natural language processing tasks by testing a GPT-2 Base model on the WikiText-2 dataset and a BERT model on the GLUE benchmark validation set \cite{wang2018glue}.
As shown in Table~\ref{tab:af_comparison}, DAPA(16) approaches perform well, achieving a maximum perplexity of 29.47 on WikiText-2 even after quantization, very close to the full-precision baseline. They also exhibit a better performance than ~\cite{wang2024improving}, where a perplexity of 31.02 is reported.
However, the MSE-based approximation attains only a PPL of 36.50, indicating a substantial performance degradation compared to the DWMSE-guided design.

\begin{table*}[htbp]
\small
\begin{center}
\begin{tabular}{lcccccccc}
\toprule
\textbf{Method} & \textbf{SST-2} & \textbf{MRPC (F1)} & \textbf{STS-B} & \textbf{QQP} & \textbf{MNLI-m} & \textbf{MNLI-mm} & \textbf{QNLI} & \textbf{Average} \\
\midrule
Baseline in \cite{peltekis2024reusing} & 91.51 & 88.01 & 88.32 & 90.77 & 83.83 & 84.43 & 90.88 & 88.24 \\
\cite{peltekis2024reusing} & 91.40 & 88.05 & 88.27 & 90.76 & 83.77 & 84.33 & 90.88 & 88.21 \\
\cite{wu2025low} & 91.86 & 88.22 & 88.30 & 90.62 & 83.82 & 84.17 & 90.68 & 88.24 \\
\midrule
Baseline in \cite{wang2024improving}& 92.08 & 85.66 & - & 91.35 & 84.23 & - & 91.49 & 88.96 \\
\cite{wang2024improving} & 91.96 & 86.27 & - & 91.12 & 84.10 & - & 91.23 & 88.94\\
\midrule
\textbf{PyTorch, F}          & 92.43 & 91.35 & 88.05 & 90.91 & 84.57 & 84.45 & 91.54 & 89.05 \\
\textbf{D(16), S\&A, F}       & 92.09 & 91.32 & 88.10 & 90.99 & 84.75 & 84.71 & 91.62 & 89.08 \\
\textbf{D(16), S\&A, Q(9.4)}  & 91.86 & 91.33 & 88.06 & 90.78 & 84.76 & 84.48 & 91.29 & 88.94 \\
\bottomrule
\multicolumn{9}{l}{\textit{*Owing to variations in testing methodologies and model checkpoints, BERT baseline results often differ. To ensure a fair}} \\
\multicolumn{9}{l}{\textit{comparison, we evaluate methods based on the performance change relative to their respective reported baselines.}} \\
\end{tabular}
\label{tab:bert_results}
\end{center}
\end{table*}

Table~\ref{tab:bert_results} summarizes the GLUE results. As different works report slightly varing baseline scores due to variations in checkpoints, fine-tuning and hyperparameters, we explicitly list the baselines used in prior work to ensure a fair comparison. Our PyTorch FP32 run (i.e., (PyTorch, F)) serves as \emph{our own} reference baseline for all DAPA experiments.  
In this context, applying DAPA(16) to both softmax and GELU in full precision (i.e., (D(16), S\&A, F)) delivers good performance, yielding a higher average score of 89.06. After applying quantization (i.e., (D(16), S\&A, Q9.4)), the average score remains at 88.94, showing a very small amount of difference from the baseline. It is worth noting that our approximation scheme, combined with quantization, is more hardware-friendly, significantly reducing power and latency while improving area efficiency, which will be elaborated on later.

\subsection{Training Transformers with DAPA} 
\begin{figure}[b]
  \centering
  \includegraphics[width=0.5\textwidth]
  {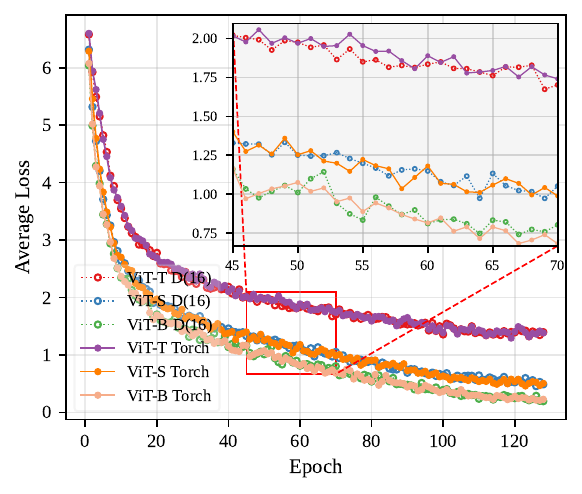}
  \caption{ Training loss curves for three ViT variants comparing standard PyTorch GELU activation with the DAPA(16) approximation function across epochs.}
  \label{fig:training}
\end{figure}

The DAPA function is designed to approximate both activation functions in the forward process and their derivatives in the backpropagation. Therefore, it can be applied for on-device fine-tuning as well as training from scratch. We thereby train ViT-Tiny, ViT-Small, and ViT-Base models from scratch using DAPA-based functions. As Figure~\ref{fig:training} illustrates, the models exhibit convergence speeds comparable to those of the baseline models. 
Moreover, as shown in Table~\ref{tab:training}, our method achieves even higher final accuracy compared to the baseline. Specifically, ViT-Small trained with DAPA(16) obtains an accuracy of 68.35\%, showing an accuracy improvement of 0.65\% over the GELU baseline. 
These results reveal that DAPA is inherently efficient for Transformer model training and fine-tuning.

\begin{table}[htbp]
 
\caption{  Classification results on ImageNet-1K: ViT models trained with GELU and DAPA activation functions}
\begin{center}
\begin{tabular}{llll}
\toprule
\textbf{} & \textbf{ViT-Base} & \textbf{ViT-Small} & \textbf{ViT-Tiny} \\
\midrule
GELU (PyTorch) & 69.76\% & 67.70\% & 66.27\% \\
DAPA(16) & 69.90\% & 68.35\% & 66.47\% \\
\bottomrule
\hline
\end{tabular}
\label{tab:training}
\end{center}
\end{table}

\subsection{Hardware Implementation} 
\begin{figure}
  \centering
  \includegraphics[width=0.6\textwidth]{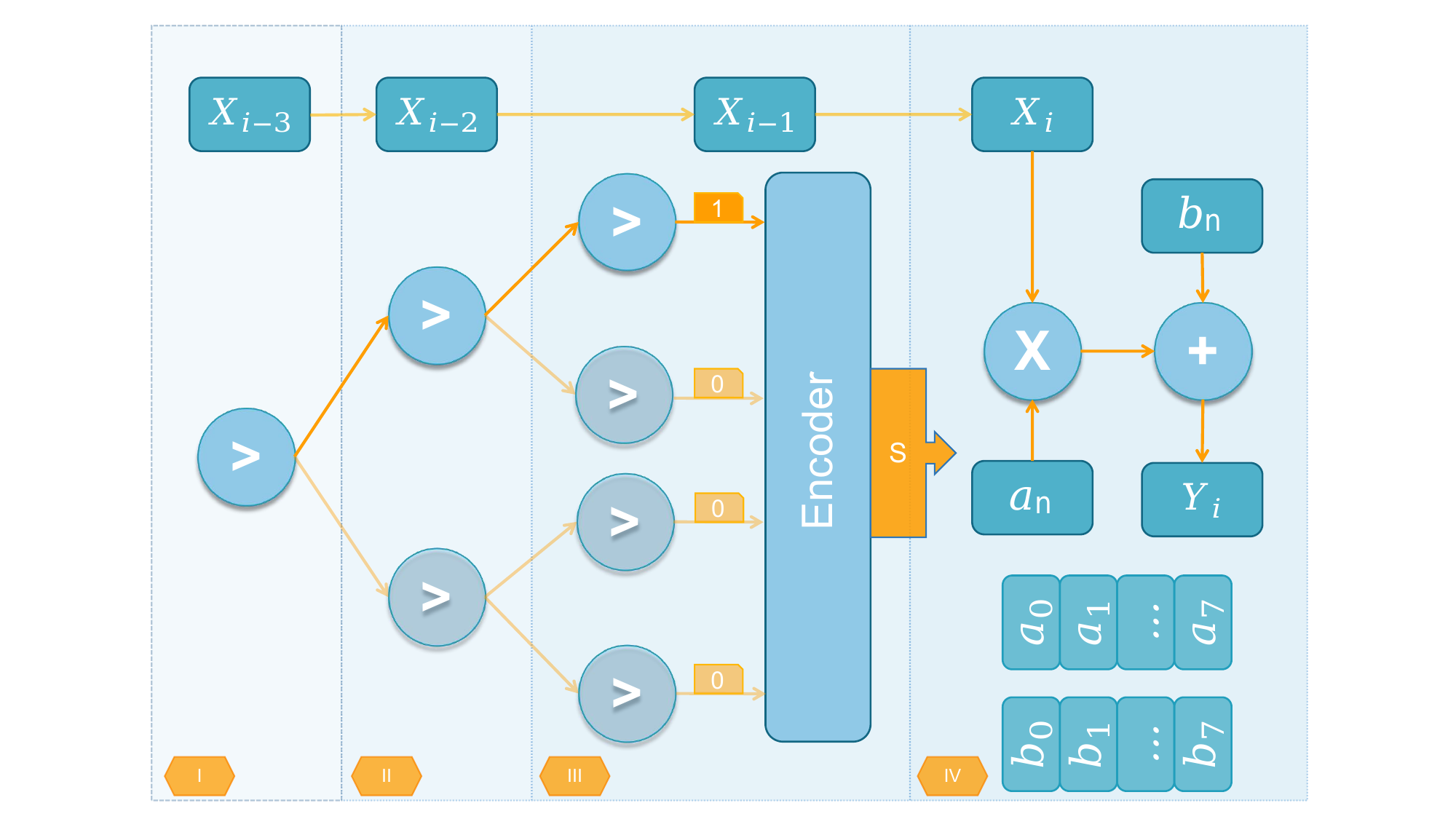}
  \caption{   Hardware architecture of the proposed DAPA with $N{=}8$ segments with four-stage pipeline. Stages I–III realize a pipelined comparator tree and encoder for segment identification, while stage IV performs the final linear computation with a single MAC unit using the selected $(a_n, b_n)$.}
  \label{fig:dapa_hardware}
\end{figure}

\begin{table}[htbp]
 
\caption{  Comparison of FPGA Hardware Implementations for Activation Functions.}
\begin{center}
\begin{tabular}{llccccc}
\toprule
& \textbf{DType$^{\mathrm{a}}$} & \textbf{L(ns)$^{\mathrm{b}}$} & \textbf{B$^{\mathrm{c}}$} & \textbf{DSP} & \textbf{FF} & \textbf{LUT} \\
\midrule
Gelu (Ours)&FP32 & 580 & 5 & 62 & 12158 & 7228 \\
Gelu (Ours)&Fix16 & 320 & - & 8 & 5858 & 6302 \\
D(16) (Ours)$^{\mathrm{d}}$ &FP32 & 150 & 2 & 7 & 1304 & 1100    \\
D(16) (Ours)$^{\mathrm{d}}$ & Fix16 & 20 & - & 1 & 100 & 401      \\
DS(16) (Ours)$^{\mathrm{e}}$ & Fix16 & 155 & - & 1 & 2243 & 1614   \\
Gelu    \cite{sadeghi2024peano}  & Fix16 & - & - &16 & 2951 & 2940      \\
Softmax \cite{sadeghi2024peano}  & Fix16 & - & - &48 & 3831 & 5595      \\
Gelu    \cite{bai2023ltrans} & FP32 & - & - &0 & 2499 & 11314       \\
Softmax \cite{bai2023ltrans} & FP32 &  - & - &0 & 13837 & 238569  \\
G\&S    \cite{wu2025low}$^{\mathrm{d}}$ & Fix16 & - & - &0 & 1024 & 4897 \\
\bottomrule
\multicolumn{7}{l}{$^{\mathrm{a}}$Data Type;  $^{\mathrm{b}}$Latency;  $^{\mathrm{c}}$BRAM;} \\
\multicolumn{7}{l}{$^{\mathrm{e}}$Integrating DAPA into the Full Softmax Unit;} \\
\multicolumn{7}{l}{$^{\mathrm{d}}$Reconfigurable to approximate GELU or Softmax Exp;} \\
\end{tabular}
\label{tab:hardware_comparison_full_detailed}
\end{center}
\end{table}

We design an efficient DAPA hardware engine for edge deployment. Fig.~\ref{fig:dapa_hardware} illustrates an 8-segment (i.e., $N=8$), four-stage pipeline, where a $\log_2 N$-stage comparator tree selects the segment index $n$, and a LUT provides $(a_n,b_n)$ to a single DSP to compute $y = a_n x + b_n$. By replacing the original nonlinearity with a few comparisons and one MAC unit, the architecture significantly reduces latency and resource utilization compared to conventional activation blocks.

Specifically, we implement $ \tanh$-based GELU, aligned with our PyTorch version, in both 32-bit Floating Point format (FP32) and 16-bit quantized fixed-point format (Fix16), and summarize the results in Table~\ref{tab:hardware_comparison_full_detailed}. Our FP32 Gelu is implemented using the PyTorch $\tanh$ approximation with HLS. Our DAPA(16) engine is synthesized with Vitis HLS~2025.1 at 200MHz. It is aligned with the (D, S\&A, Q) configuration and achieves network performance comparable to the FP32 baseline.
The FP32 reconfigurable DAPA(16) unit (for both GELU and Exponential computation) achieves a latency of 150 ns using only 7 DSPs, 1304 FFs, and 1100 LUTs, while the Fix16 DAPA(16) core achieves a lower latency of 20ns using a single DSP, 100 FFs and 401 LUTs. Compared with prior Fix16 GELU block~\cite{sadeghi2024peano}, DAPA(16) reduces DSP usage by $16\times$ and achieves one order of magnitude savings in flip-flops and LUT resources.

For a fair comparison with the benchmark design, we enhance the DAPA unit by incorporating additional accumulators and a divisor-equivalent unit to form a fully-functional Softmax module, denoted as $DS(16)$. Our $DS(16)$ unit utilizes 2243 FFs and 1614 LUTs, both of which are lower than the figures reported in \cite{sadeghi2024peano}. More importantly, it achieves a 48-fold reduction in DSP usage compared to \cite{sadeghi2024peano} with a latency of 155 ns.
\section*{Conclusion}
We have introduced DAPA as an efficient, hardware-friendly approach to approximate activation functions for Transformer models. By exploiting input distributions, DAPA enables efficient approximation of activation functions and their derivatives without compromising performance. Together with the proposed DWMSE loss function and the DWMSE-guided fixed-point quantization, our scheme provides a software-hardware co-design methodology that enhances approximation performance while reducing hardware cost. We further show that DAPA is fully trainable with a comparable convergence rate and thus well-suited for both on-chip inference and training. These results highlight the promise of distribution-aware activation functions for future Transformer accelerator design. 

\bibliographystyle{unsrt}  
\bibliography{references}  

\end{document}